\documentclass{article}
\usepackage{amsmath,amsfonts,amsthm}
\usepackage{algorithmic}
\usepackage{algorithm}
\usepackage{array}
\usepackage[caption=false,font=normalsize,labelfont=sf,textfont=sf]{subfig}
\usepackage{textcomp}
\usepackage{stfloats}
\usepackage{url}
\usepackage{verbatim}
\usepackage{graphicx}
\usepackage{cite}
\usepackage{ifthen}
\usepackage{intcalc}
\usepackage{multirow}
\usepackage{xcolor}
\usepackage{environ}
\usepackage{tikz}
\usepackage{xspace}

\usetikzlibrary{math, calc, decorations.pathmorphing}
\newtheorem{definition}{Definition}

\newcommand{\wetsand}{WETSAND\xspace}

\NewEnviron{toremove}{\color{red}\BODY\color{black}}

\begin{document}
\title{Warped Time Series Anomaly Detection}

\author{Charlotte Lacoquelle\textsuperscript{1},
Xavier Pucel\textsuperscript{2},
Louise Travé Massuyès\textsuperscript{3},\\
Axel Reymonet\textsuperscript{1}, and
Benoît Enaux\textsuperscript{1} \\
\\
\textsuperscript{1}Vitesco Technologies, Toulouse, France\\
\textsuperscript{2}ONERA, DTIS, Université de Toulouse, France \\
\textsuperscript{3}LAAS-CNRS, Université de Toulouse, CNRS, France\\
}


{Lacoquelle et al. Warped Time Series Anomaly Detection}


\maketitle

\begin{abstract}
This paper addresses the problem of detecting time series outliers, focusing on systems with repetitive behavior, such as industrial robots operating on production lines.
Notable challenges arise from the fact that a task performed multiple times may exhibit different duration in each repetition and that the time series reported by the sensors are irregularly sampled because of data gaps. The anomaly detection approach presented in this paper consists of three stages.
The first stage identifies the repetitive cycles in the lengthy time series and segments them into individual time series corresponding to one task cycle, while accounting for possible temporal distortions.
The second stage computes a prototype for the cycles using a GPU-based barycenter  algorithm, specifically tailored for very large time series.
The third stage uses the prototype to detect abnormal cycles by computing an anomaly score for each cycle.
The overall approach, named WarpEd Time Series ANomaly Detection (\wetsand), makes use of the Dynamic Time Warping algorithm and its variants because they are suited to the distorted nature of the time series.
The experiments show that \wetsand scales to large signals, computes human-friendly prototypes, works with very little data, and outperforms some general purpose anomaly detection approaches such as autoencoders.
\end{abstract}


\section{Introduction}
\label{introduction}

The widespread expansion of connected devices and sensors in any domain (medical, financial, industrial) results in a massive production of data streams - where values are ordered or explicitly timestamped. This creates opportunities for automatically monitoring the execution of tasks in order to detect any deviation from normal execution, and react accordingly. 

The quality of the data produced by sensors and data collectors can significantly affect the performance of monitoring approaches. In particular, we are interested in detecting abnormal cycles inside time series collected during the execution of repetitive tasks. In this setting, the considerable variation in cycle time and the need to accommodate missing data pose considerable challenges.

While this work applies to any dataset of cyclostationary time series, it is initially motivated by the need of monitoring robotic arms that operate on production lines. While these are equipped with inner fault detection mechanisms, they fail to detect problems related to their environment (setup, calibration, programming, product defects, etc). Our objective is to detect improper execution of repetitive tasks by the robots through by detecting abnormal execution cycles in their monitoring data. Several factors make it difficult to reuse the state of the art. First, the cycles are very long (in the tens of thousands of time points), and it has been observed that downsampling the time series degrades the detection quality. Second, the cycles may vary in length, because the robot program contains physical tasks of variable duration (eg object gripping, bar code scans), or simply because the operator takes a manual action. Operator intervention can be as limited as pausing and resuming the whole production line, or as intrusive as moving the robot in a different position, skipping part of its task, or re-calibrating it in the middle of a cycle. Ideally, the former interventions should not be detected as abnormal situations, while the latter should. Finally, we aim at being robust to a small but significant amount of data gaps in the acquisition process. This phenomenon produces inter-series and intra-series irregularities \cite{sun2021te}. Inter-series irregularities refer to deviations that occur between different time series. In contrast, intra-series irregularities involve deviations within a single time series.

\subsection{Scientific problem}
\label{ssec:scient_problem}
This paper addresses the problem of detecting outlier time series inside a set of time series, in the context of intra-series irregularities. It targets systems exhibiting repetitive behavior and generating cyclostationary time series through continuous monitoring. The challenges posed are as follows: 
\begin{itemize}
\item The time series represent cycles extracted from cyclic operation trajectories. They are ideally periodic, except that the irregularities mentioned above make them quasiperiodic (as defined in \cite{liu2020anomaly}). Given that the global shape of abnormal trajectories is not affected, the use of clustering would not be efficient.
\item There are no labeled examples of what is considered normal behavior, therefore an unsupervised approach is required.
\item The anomalies manifest themselves as gradual deviations in the time series rather than as isolated point outliers.
\item The large gaps of missing data defeat window-based approaches, emphasizing the need to study the cycle signal as a whole.
\end{itemize}
The main contributions of this work are summarized as follows:
\begin{itemize}
    \item A segmentation algorithm based on the Subsequence Dynamic Time Warming (Subsequence-DTW) algorithm that enables segmentation of a long sequence of repetitive task data in cycle sequences despite temporal warping.
    \item A novel GPU version of the Soft-DTW barycenter algorithm with chunk-splitting, which makes possible to obtain in reasonable time a representative normal cycle as the barycenter of a training set of cycle sequences.
    \item A time series outlier detection algorithm, named WarpEd Times Series ANomaly Detection (\wetsand), that allows warping in the time series similarity scoring used to detect anomalies 
    \item A wide range of experiments involving different parameter settings, comparing results with competing algorithms in the field of Deep Learning, and validating the approach on real industrial robot data.
    
\end{itemize}

\subsection{Motivating use case}
\label{Motivating_usecase}

The work reported in this paper is motivated by a plant that manufactures printed circuit boards. It features a fleet of 6-axis industrial robots by Universal Robots. Each robot is programmed for a specific task such as palletizing, stacking, or holding a camera for quality control. The repetitive nature of their programmed task causes the output time series to be, in theory, cyclic. However, some physical interactions, such as scanning a bar code or grasping an object, have variable duration, and as each cycle contains several dozens of such interactions, its length varies significantly in practice.

All the robots in the plant are connected to a monitor that collects the timestamp, the angular position of the 6 robot joints, the pose of the robot's end effector, and other variables such as the current and voltage measurements of the joint motors.
The monitoring data is sampled at a frequency of 125 Hz. For various design, implementation and deployment reasons, data may be lost during periods of time of varying length, at irregular intervals. In other words, sampling is very irregular, and the time series feature large data gaps. This impacts diagnosis as these gaps contain complex patterns that cannot be reconstructed via interpolation methods. Prediction approaches also fail to impute the missing data, because in addition to the missing data, the physical robot cycles also contain pauses, vary in length, and the interactions with data gaps impair the accuracy of predictors.

While our approach applies to any robot or even any system that performs cyclic tasks, this paper focuses on a robot that fills containers with electronic devices at the end of a production line.
Note that the health state of the robot is unknown and it is not possible to reproduce a laboratory experiment that could provide a reference behavior for a cycle as in \cite{bittencourt2012data}. 

The paper is organized as follows. Section~\ref{SOTA} provides a state of the art of time-series anomaly detection and elastic distance measures. Section~\ref{Algo_design} provides background material on Soft-DTW and describes the \wetsand approach with a novel implementation of Soft-DTW. Section~\ref{Evalu_approach} describes the use case and the dataset on which the approach is evaluated. Section~\ref{Experiments} evaluates the performances of \wetsand on the use case and provides a discussion on the algorithm's fine tuning. Finally, Section~\ref{Conclu} concludes the paper.

\section{State of the Art}
\label{SOTA}

The analysis of systems that perform repetitive tasks has already been addressed in various manners. In \cite{bittencourt2012data}, the authors exploit the repetitive behavior of a robot - it could be any automated system - to generate fault indicators. They rely on nominal data, the first execution of the robot task, to estimate the joints condition, by monitoring the changes in the distribution of data batches. In \cite{liu2020anomaly}, a two stage approach is used, where a time series is first segmented into quasi-periods, that are fed to an anomaly detection algorithm based on a deep neural network featuring recurrent and convolutional layers. Time series segmentation can also be performed using Dynamic Time Warping (DTW) \cite{sakoe1978}, and more precisely its variant Subsequence-DTW \cite{sakurai2006stream} as in \cite{barth_2013}, to identify patterns by looking for local minimums that are under a fixed threshold. In \cite{lacoquelle2022deviation}, segmentation is also achieved through a Subsequence-DTW variant, before anomaly detection is performed with DTW. This approach relies on a reference trajectory produced by a high-fidelity simulator, which is costly to produce.

In time series, anomalies or outliers refer to values or patterns that do not conform with the expected behavior \cite{aggarwal2017}. In \cite{blazquez2021review} the authors present a taxonomy of outlier detection techniques in time series data. They distinguish three types of outliers: point outliers, subsequence outliers and outlier time series. The latter refers to a time series that differs in its entirety from the rest of the multivariate time series population. 

The work of this paper falls within the detection of outlier time series. The approach is unsupervised because in our context, labels are rarely available.


AutoEncoder approaches are quite widespread, where a neural network is trained to reconstruct the system's normal behaviour, and poor reconstruction quality is interpreted as an anomaly. \cite{kieu2019outlier} uses an autoencoder with dilated convolution and temporal mechanisms to electrocardiogram data. \cite{xu2018unsupervised} was one of the first papers to introduce a proper theoretical explanation for anomaly detection with a VAE on time series. Since then, the detection accuracy of VAE models has been improved by introducing discriminators to which are fed the raw and reconstructed data \cite{vaegan_bearing}.


Many of the aforementioned approaches use deep neural networks (DNNs) that suffer from a long training time, laborious hyper-parameter tuning, and require high computational resources. Thus, as the study \cite{audibert2022deep} demonstrates, before choosing complex DNNs, and especially in the context of real-world applications, it is crucial to evaluate whether conventional methods are more appropriate.

Irregularly sampled data can prevent many techniques from performing well, and is an issue that must be addressed specifically. For instance, in \cite{li2015classification}, a classification task is possible despite very sparsely sampled datasets, enabled by the Gaussian process representation. In \cite{sun2021te}, the authors propose a time encoding mechanism that models the irregularities in the signal, and train a Recurrent Neural Network (RNN) for a prediction task that usually requires a uniform time distribution

Another approach to handle missing data is to ignore missing data points and compare time series through elastic distances.
%
%
\cite{paparrizos2020debunking} performs a comprehensive evaluation of 71 time series distance measures but the design of new measures is an active research field.
The Dynamic Time Warping (DTW) metric \cite{sakoe1978} is well known as an appropriate measure of similarity between time series because it handles time deformation and irregular sampling.
K-Shape is much faster than Dynamic Time Warping (DTW) \cite{sakoe1978}, however it only accounts for shifting the time series, and not for warping it.
The DTW metric has been widely used for analysing time series, including for detecting outlier time series. In \cite{zhang2022predicting}, DTW is used as the distance of the k-means clustering algorithm. However, anomaly detection itself is then performed inside each cluster using a decision tree approach.  

Some elastic distances offer ``averaging'' capabilities, the average time series being the one that minimizes the value of some elastic distance to a set of time series. This is useful for computing a barycenter (also known as centroïd, pseudo-average, prototype or consensus object) that represents a set of time series. In \cite{petitjean2012summarizing} DTW is extended to a set of N signals, and a genetic algorithm is used to find a signal that minimizes its value. DTW Barycenter Averaging (DBA) \cite{petitjean2011global} is a popular algorithm that provides a prototype through a subgradient iterative method. The more recent Soft-DTW formulation of DTW presented in \cite{cuturi2017soft} presents a smoothed DTW that uses the soft-minimum which makes it differentiable with respect to its inputs. Soft-DTW has several applications: clustering, prototype learning for a set of time series, or time series prediction. Similarly, \cite{janati2022averaging} aims at designing the barycenter of spatio-temporal datasets; the temporal shifts are captured by Soft-DTW and the space and size invariances are handled with Unbalanced Optimal Transport.

Such barycenters can naturally be used for anomaly detection, by using the elastic distance between a time series and the barycenter of a given reference set.

\section{Warped time series anomaly detection with \wetsand} \label{Algo_design}

The method presented in this paper, named \wetsand (from WarpEd Time Series ANomaly Detection), takes as input quasiperiodic time series with missing data. It is organized in three steps that will be called \emph{Step 1}, \emph{Step 2} and \emph{Step 3} in the paper:
\begin{enumerate}
    \item \emph{Segmentation into cycles}: starting with a reference cycle (with possibly missing data), we use the Subsequence-DTW (Sub-DTW) algorithm to segment the data stream into quasi periods, or cycles. This stage uses all the signal dimensions together to better accommodate for the data gaps that happen irregularly across the cycles.
    \item \emph{Prototype computation}: a set of cycles is used to obtain a prototype cycle with the Soft-DTW barycenter algorithm \cite{cuturi2017soft}. As available implementations fail to scale up to tens of thousands of time steps, we describe a novel GPU implementation specifically tailored to long time series.
    \item \emph{Outlier detection}: outlier cycles are detected by checking their similarity with the prototype cycle according to the DTW metric. Threshold-based approaches provides satisfactory performance for anomaly detection, more elaborate approaches are left for future work.
\end{enumerate}

After introducing a few notations, this section describes each step in detail.

\subsection{Modeling hypotheses and notations} \label{modeling}
The input of the \wetsand approach is a multivariate time series 
$\mathbb{Y} = [y_1,\ldots,y_K]$
of length \(K \in  \mathbb{N} \), where $y_i, i=1,\dots,K$ are d-dimensional, $d \in (\mathbb{N}\cup\{nan\})^*$, column vectors acquired with a theoretical sampling rate $r$. The $nan$ values represent time steps with missing values, however DTW and its variants do not support them. As a consequence, we simply remove them from the regularly sampled time series $\mathbb{Y}$ and obtain an irregularly sampled time series $Y \in \mathbb{R}^{d \times N}$:
$Y = [y_1,...,y_N]$
of length \(N \leq K\).

The time series $Y$ can be several hours long with repeated patterns; a repetition is called a \emph{cycle trajectory}. As explained in section \ref{Motivating_usecase}, the length of each cycle trajectory can significantly vary.

We assume that a reference trajectory $X_{ref} \in \mathbb{R}^{d \times M}$ is provided by an expert. $X_{ref}$ describes a single irregularly sampled cycle, and we assume that $M \ll N$.
We have deliberately opted for an approach where an expert manually defines the boundaries of the reference cycle. The reference being the initial input of the method, this ensures that the results of the three steps of \wetsand remain easily interpretable for the expert involved.



In a first stage, the data stream  is segmented into a set of cycle trajectories that match the reference trajectory. The segments are gathered into a set $C_{traj}$, and used to compute a prototype $B$.

In a second stage, the incoming data stream is also segmented into a set of cycle trajectories that match the barycenter $B$. The cycles are gathered into a set $C_{test}$ upon which anomaly detection is performed.

In this paper, for practical reasons and without loss of representativity, a single data stream $Y\in\mathbb{R}^{d\times N}$ is gathered and segmented into cycles that match a reference trajectory $X\in\mathbb{R}^{d\times M}$. The identified cycles are gathered into a set $C$ that is split into $C_{train}$ and $C_{test}$.

\begin{figure}
    \centering
    \includegraphics[width=\linewidth]{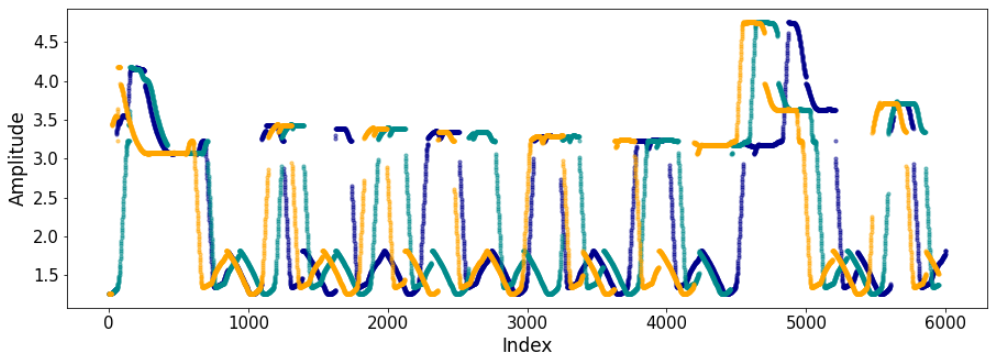}
    \caption{Three trajectories corresponding to three repetitions of the same task, with missing data and time warping that results from the variable duration of physical interactions (handling, sensing).}
    \label{fig: timeseries}
\end{figure}

The 3 core functions of \wetsand are now defined in order: segmentation into cycles, prototype computation, and anomaly detection.

\subsection{Segmentation into cycles} \label{Sub_DTW}

The first function of \wetsand uses Subsequence-DTW (Sub-DTW), a pattern-recognition technique allowing temporal deformations\cite{sakurai2006stream}. This section introduces definitions for the DTW and the Sub-DTW metrics.

\begin{definition}[Alignment path]
    Given two multivariate time series of dimension $d$ represented by \(X = [x_1,...,x_M] \in \mathbb{R}^{d \times M}\) and \(Y = [y_1, ..., y_N] \in \mathbb{R}^{d \times N}\), an \emph{Alignment path} $\pi$ between $X$ and $Y$ is a sequence of pairs of indices $\pi = [(i_1,j_1), \ldots, (i_k, j_k)]$ such that two consecutive pairs $(i,j)$ and $(i^\prime,j^\prime)$ satisfy:
    \begin{align*}    
    (i^\prime, j^\prime) \in \big\{(i, j+1), (i+1, j), (i+1, j+1)\big\}
    \end{align*}
     A \emph{subsequence} alignment path is an alignment path such that $i_1=1$ and $i_k=M$. 
    A \emph{complete} alignment path is such that $i_1 = j_1 = 1$, $i_k = M$ and $j_k=N$, i.e. it starts at $(1, 1)$ and ends at $(M, N)$.
\end{definition}

\begin{definition}[Alignment matrix]
    An alignment path between two time series \(X = [x_1,...,x_M] \in \mathbb{R}^{d \times M}\) and \(Y = [y_1, ..., y_N] \in \mathbb{R}^{d \times N}\) can be represented as an \emph{Alignment matrix} $A \in \{0, 1\}^{M \times N}$ that contains $1$ at the index pairs that belong to the alignment path and $0$ elsewhere.
    
    The set of all alignment matrices for \emph{complete} alignment paths of size $M \times N$ is denoted $\mathcal{A}_{M,N}$.

    The set of all alignment matrices for \emph{subsequence} alignment paths of size $M \times N$ is denoted $\mathcal{A}^{sub}_{M,N}$.
\end{definition}

\begin{definition}[Distance Matrix]
    Given two time series \(X = [x_1,...,x_M] \in \mathbb{R}^{d \times M}\) and \(Y = [y_1, ..., y_N] \in \mathbb{R}^{d \times N}\), and a distance function $\delta : \mathbb{R}^{2 \times d \times M} \to \mathbb{R}$, the distance matrix $\Delta(X,Y) \in \mathbb{R}^{M \times N}$ such that $\Delta(X,Y)_{i,j} = \delta(x_i, y_j)$.
\end{definition}
The distance function $\delta$ is taken as the Euclidean distance, or $l^2$ norm. 


\begin{definition}[DTW]
    Given two time series \(X = [x_1,...x_M] \in \mathbb{R}^{d \times M}\) and \(Y = [y_1,...,y_N] \in \mathbb{R}^{d \times N}\), the cost of each alignment path $\pi_{\alpha}$ is equal to the inner product of its alignment matrix $A_{M,N}$ with the distance matrix $\Delta(X,Y)$.
    
    The DTW distance between $X$ and $Y$ is defined as the minimal alignment cost among all possible complete alignment paths between $X$ and $Y$ as follows:
    \begin{equation} \label{alignmt_cost_dtw} 
        DTW(X, Y) = \min_{A \in \mathcal {A}_{M,N}}\langle A,\Delta (X, Y)\rangle 
    \end{equation}
\end{definition}

In practice, the optimal alignment path and its associated alignment cost are obtained through a dynamic programming algorithm \cite{sakoe1978}.
Sub-DTW is a variant of DTW that considers subsequence alignment paths. It is defined by:
\begin{align}
\begin{split}
    \textit{Sub-DTW}(X,Y) & = \min_{A \in \mathcal {A}^{sub}_{M,N}}\langle A,\Delta (X, Y)\rangle \\
    & = \min_{0 \leq a \leq b \leq N} DTW(X, Y[a:b])
\end{split}
\end{align}
Where $Y[a:b] = [y_a, \ldots, y_b]$ is the subsequence of $Y$ bet\-ween indices $a$ and $b$. Intuitively, Sub-DTW matches the whole time series $X$ against the best subsequence of $Y$, usually noted $Y[a^*:b^*]$.

With Sub-DTW defined, the \wetsand segmentation algorithm of a time series $Y = [y_1, \ldots y_N] \in \mathbb{R}^{d\times N}$ with a reference quasi-period $X \in \mathbb{R}^{d\times M}$ is the following:
\begin{enumerate}
    \item Initialize the set $C_{traj}$ as the empty set and the sliding window indices at $ws=0, we=2M$.
    \item Compute $\textit{Sub-DTW(X, Y[ws:we])}$ and note $a^*$ and $b^*$ the first and last indices along $Y$ in the subsequence alignment path.
    \item Add the time series $Y[a^*:b^*]$ to the set $C_{traj}$.
    \item Update the sliding window indices to $ws=b^*-\alpha$, $we=ws+2M$.
    \item While the sliding window overlaps $Y$, i.e. $ws < N$, loop back to step 2).
\end{enumerate}
$\alpha$ is a constant set at $0.15M$ in order to ensure that missing data between two cycles does not interfere with the cycle segmentation.
When the algorithm terminates, $C_{traj}$ contains the cycles in $Y$ that were matched with $X$ by the Sub-DTW algorithm.
The segmentation step of \wetsand is illustrated in the evaluation section, in Figure~\ref{fig: segmentation_j0}.

In practice, there is a possibility that a fragment of the time series $Y$ (between indices $ws$ and $a^*$ at step 4) is not part of any cycle, and discarded. While we did not encounter this situation in our experiments, it does not discredits our approach: at the prototype computation stage these data can simply be discarded; at the anomaly detection stage, this unassigned data constitutes in itself an anomaly. The anomaly detection stage can focus on identified cycles to detect more subtle anomalies.

\subsection{Prototype computation} \label{Soft_DTW}

The second function of \wetsand relies on the Soft-DTW barycenter algorithm. Soft-DTW is another variant of DTW, first described in \cite{cuturi2017soft}, that achieves a soft-minimum of all alignment costs. The principle behind the Soft-DTW variant is to replace the $\min$ operator with a soft $\min^\gamma$ operator defined as follows:
\begin{equation} \label{min_operator}
    {\min}^\gamma (a_{1},\ldots,a_{n}) =
    \begin{cases}
        \displaystyle \min\nolimits_{i=1}^n a_{i} & \text{if }\gamma = 0 \\
        \displaystyle -\gamma \log \sum\nolimits_{i=1}^{n}e^{-a_{i}\gamma} & \text{if } \gamma >0
    \end{cases}
\end{equation}
where $a_{1}, \ldots, a_{n}$ are the respective costs of the alignment paths $\pi_{1}, \ldots, \pi_{n}$ and $\gamma$ is a smoothing parameter. 

\begin{definition}[Soft-DTW]
    Given two multivariate time series \(X = [x_1,...x_M] \in \mathbb{R}^{d \times M}\) and \(Y = [y_1,...,y_N] \in \mathbb{R}^{d \times N}\),, the Soft-DTW distance between $X$ and $Y$ is computed like $DTW(X,Y)$ in (\ref{alignmt_cost_dtw}), but using the soft $\min^\gamma$ operator instead of $\min$:
    \begin{align} \label{soft_dtw}
        \text{Soft-DTW}^\gamma(X,Y) = \mathop{\min\nolimits^\gamma}\limits_{A\in \mathcal {A}_{M,N}}\big(\langle A, \Delta (X,Y)\rangle\big)
    \end{align}
\end{definition}

The dynamic programming algorithm that computing Soft-DTW is similar to that of DTW, and consists in computing the cumulative cost matrix.

\begin{definition}[Cumulative alignment cost matrix]
    Given two multivariate time series \(X = [x_1,...x_M] \in \mathbb{R}^{d \times M}\) and \(Y = [y_1,...,y_N] \in \mathbb{R}^{d \times N}\), $\textit{Soft-DTW}^\gamma(X,Y)$ is computed as follows:
    \begin{align}
    \begin{split}
    &r_{i,j} = \left\{\begin{array}{l}
         0 \hfill \text{if } i = j = 0\\
         \infty \hfill \text{if } i = 0 \text{ or } j = 0\\
         \delta_{i,j} + min^\gamma\left(\begin{array}{c}
         r_{i-1, j-1},\\r_{i-1, j},\\ r_{i, j-1}\end{array}\right) \text{otherwise}
    \end{array}\right.\\
    &\textit{Soft-DTW}^\gamma(X,Y) = r_{M,N}
    \end{split}
    \label{eq-r-matrix}
    \end{align}
    Where $\delta_{i,j}$ is shorthand for $\Delta(X,Y)_{i,j}$.
    The cumulative alignment cost matrix $R(X,Y)$ of size $M\times N$ retains the values $r_{i,j}$ for $i > 0$ and $j  > 0$.
\end{definition}

\begin{figure}
    \centering
    \includegraphics[width=\linewidth]{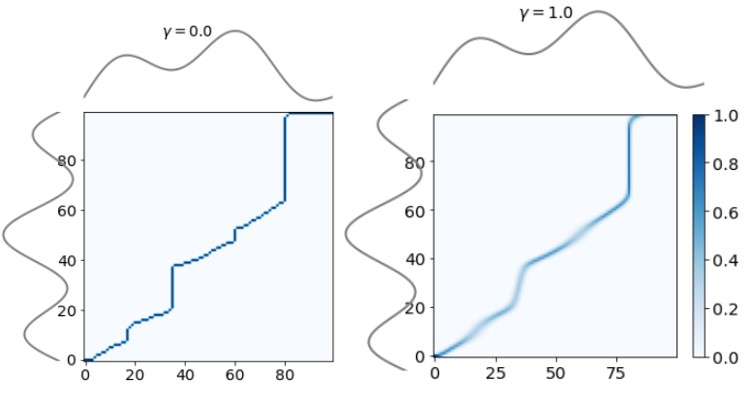}
    \caption{Two time series X and Y and the optimal warping path according to DTW (left) and according to Soft-DTW (right), illustrated through the alignment matrix (graph obtained using Python matplotlib library).}
    \label{fig: soft_dtw_matrices}
\end{figure}

Figure \ref{fig: soft_dtw_matrices} depicts the optimal soft alignment obtained with $\gamma = 0$ which is equivalent to the classical DTW alignment, and the alignment obtained with $\gamma = 1$.

The main advantage of Soft-DTW with respect to the standard DTW is that it is differentiable everywhere with respect to its input signal. 

The partial derivative of $\textit{Soft-DTW}(X, Y)$ with respect to $X$ is a vector defined as follows:
\begin{equation}
\nabla_{X}\,\textit{Soft-DTW}^{\gamma}(X,Y) =
    \left(\frac{\partial \Delta(X, Y)}{\partial X} \right)^T E,
    \label{eq:eq-gradient}
\end{equation}
where $\Delta(X,Y)$ is the distance matrix between $X$ and $Y$, and $E$ is a matrix that can be computed by a dynamic programming algorithm described in \cite{cuturi2017soft} that requires $O(MN)$ memory, much like the DTW and Sub-DTW computations.

We use this partial derivative to compute an approximation of the Soft-DTW barycenter of a set of signals by gradient descent in a straightforward manner.

\begin{definition}[Soft-DTW barycenter]
    Given a family of time series $Y_1, \ldots, Y_T$ of dimension $d$, of respective lengths $m_1, \ldots, m_T$ and with respective weights $\lambda_1, \ldots, \lambda_T$, a smo\-othing parameter $\gamma$ and a barycenter length $m_b$, the Soft-DTW barycenter $B$ is defined as the time series of length $m_b$ that minimizes the sum of Soft-DTW distances between $B$ and $Y_1, \ldots, Y_T$, normalized by their weights and lengths:
    \begin{equation} \label{equ_barycenter} 
        B = \arg\min_{B \in  \mathbb{R}^{d*m_b}} \sum\limits_{i=1}^T \frac{\lambda_{i}}{m_{i}} \textit{Soft-DTW}^{\gamma} (B, Y_i)
    \end{equation}
\end{definition}
Unless stated otherwise, we set the barycenter length at $m_b = \max(m_1, \ldots, m_T)$, and the weights at $\lambda_1 = \ldots = \lambda_T = 1$. 

A very efficient implementation of Soft-DTW for short time series is found in the TSlearn library \cite{TSLearn_Tavenard}. However, 
this implementation tends to be slow for long signals. For instance, computing the barycenter of 5 signals with 18385 time steps takes up to 4 hours on a cloud instance with 16 CPUs and 125GB of memory.

The barycenter $B$ computation is parallelizable to some extent: the Soft-DTW distance and its partial derivatives are completely independent for each pair $(B, Y_i), i=1, \dots,m_T$. Furthermore, each row in the $R$ and $E$ matrices can be computed in parallel with some synchronization. GPU based implementations of the Soft-DTW algorithm are described in \cite{zhu2018developing} and \cite{schmidtCuDTWUltraFastDynamic2020} but they consume too much GPU memory when applied to long signals. The remainder of this section describes an implementation focused on fitting the computation in GPU memory by splitting the $R$ and $E$ matrices into $chunks$.

The proposed implementation consists in computing the matrices $R$ of cumulative costs, and $E$ of partial derivatives as mentioned in \cite{cuturi2017soft}, by splitting them in chunks as illustrated in Figure~\ref{fig:gpu-chunks}. More precisely, the $T$-sized stack of $R$ and $E$ matrices form an $R$ and $E$ tensor of shape $\big(T, m_b, m_Y\big)$, where $T$ is the number of time series, $m_b$ the length of the barycenter, and $m_y = max(m_1, \ldots, m_T)$ is the length of the longest series in $Y$ ($m_i$ being the length of time series $i$). They are decomposed into chunks of shape $(l_T, l_b, l_Y)$, where $l_T \leq T$ is the number time series evaluated in parallel, $l_b \leq m_b$ corresponds the number of synchronized GPU threads, and $l_Y \leq max(m_1,\ldots,m_T)$ is the number of $R$ values computed by each thread in a chunk.

Tuning the chunk shape essentially depends on the GPU used and aims at enhancing parallelism, which is achieved by higher values for $l_T$ and $l_b$. Threads in the $l_b$ dimension perform synchronous computation, and current GPUs have a physical limit for this parameter. Threads in the $l_T$ do not need synchronization, so it is desirable to have the highest value for this parameter, the only limitation being the GPU memory size. Furthermore, a very small value for $l_Y$ makes the streaming mechanism (that parallelizes computation and data transfer) unbalanced and less efficient.

\begin{figure}
    \centering
    \begin{tikzpicture}[scale=0.5,>=stealth,font=\small]
        \tikzstyle{arrow}=[draw=black,line width=2pt,->,nearly opaque]
        \foreach \zc / \z in {1/.5, 1/.4, 1/.3, 0/.2, 0/.1, 0/0} {
        \begin{scope}[shift={(\z,\z)}]
            \foreach \x in {0,...,4}
            \foreach \y in {0,...,3} {
            \tikzmath{\sum = \x + \y + \zc;}
            \ifthenelse%
                {\isodd{\sum}}
                {\tikzstyle{chunk}=[fill=white];}
                {\tikzstyle{chunk}=[fill=black!30];}
                \path[chunk] (\x,\y) rectangle ++(1,1);
            }
            \draw (0,0) grid (5,4);
        \end{scope}
        }
        \draw[arrow] (0.5,3.5) -- (4.5,0.5);
        \node[anchor=center,font=\footnotesize] at (2.5,-1) {R matrix};
        \begin{scope}[shift={(7,0)}]
        \foreach \z in {.4, .3, .2} {
            \begin{scope}[shift={(\z, \z)}]
                \fill[white] (0,0) grid (1,4) rectangle (0,0);
                \fill[black!30] (0,0) rectangle (1,1);
                \fill[white!60] (0,2) rectangle (1,1);
                \fill[black!30] (0,2) rectangle (1,3);
                \draw (0,0) grid (1,4);
            \end{scope}
            \draw
            (-1.5, 4.5) edge[bend left,arrow] (0.3, 4.5)
            (1.5, 4.5) edge[bend left,arrow] (3.3, 4.5);
            }
            \node[anchor=center,font=\footnotesize] at (0.5,-1) {Chunk stack};
        \end{scope}
        \begin{scope}[shift={(10,0)}]
        \foreach \zc / \z in {1/.5, 1/.4, 1/.3, 0/.2, 0/.1, 0/0} {
        \begin{scope}[shift={(\z,\z)}]
            \foreach \x in {0,...,4}
            \foreach \y in {0,...,3} {
            \tikzmath{\sum = \x + \y + \zc;}
            \ifthenelse%
                {\isodd{\sum}}
                {\tikzstyle{chunk}=[fill=black!30];}
                {\tikzstyle{chunk}=[fill=white];}
                \path[chunk] (\x,\y) rectangle ++(1,1);
            }
            \draw (0,0) grid (5,4);
        \end{scope}
        }
        \draw[arrow] (4.5,0.5) -- (0.5,3.5);
        \node[anchor=center,font=\footnotesize] at (2.5,-1) {E matrix};
        \end{scope}
    \end{tikzpicture}
    \caption{The 3-dimensional alignment and gradient matrices are decomposed into chunks for computation. The $R$ matrix is computed in increasing time step indices, the $E$ in reverse order.}
    \label{fig:gpu-chunks}

    \vspace{\baselineskip}
    \begin{tikzpicture}[scale=0.5,>=stealth,font=\small]
        \foreach \z / \op in {0.2 / 50, 0.1 / 50, 0 / 100} {
        \begin{scope}[shift={(\z,\z)}]
        \path[draw=black!\op, fill=white] 
            (-.1,-.1) rectangle (10.1, 6.1)
            (-2, 0) rectangle (-1, 6)
            (11, 0) rectangle (12, 6)
            (0, -2) rectangle (10, -1)
            (0,  7) rectangle (10, 8)
            (-2, 7) rectangle (-1, 8)
            (11, -1) rectangle (12, -2)
            ;
        \end{scope}
        }
        \fill[black!30]
            (9.1, 0.9) rectangle (9.9, 5.9)
            (0.1, 0.1) rectangle (9.9, 0.9)
            ;
        \foreach \y/\x in {5.5/5,4.5/4,3.5/3,2.5/2,1.5/1} {
            \draw [->,decorate,decoration={snake,amplitude=.4mm,segment length=2mm,post length=1mm}]
            (0,\y) -- (\x, \y);
        }
        \draw[->] (0,4.5) -- (0.1,4.5);
        \node at (5,3) {GPU threads};
        \foreach \x in {0.5,1.5,...,9.5} {
            \draw[->] (\x, 6.9) -- (\x, 6.1);
            \draw[->] (\x, 0.3) -- (\x, -0.9);
        }
        \foreach \y in {0.5,1.5,...,5.5} {
            \draw[->] (-0.9, \y) -- (-0.1, \y);
            \draw[->] (9.7, \y) -- (10.9, \y);
        }
        \draw[->] (-0.9, 6.9) -- (-0.1, 6.1);
        \draw[->] (9.7, 0.3) -- (10.9,-0.9);
        \node[anchor=center] at (5,7.5) {Last line from upper chunk};
        \node[anchor=center,rotate=90] at (-1.5,3) {From left chunk};
        \node[anchor=center] at (5,-1.5) {To lower chunk};
        \node[anchor=center,rotate=90] at (11.5,3) {To right chunk};
    \end{tikzpicture}
    \caption{Computation of a chunk of the $R$ matrix with several synchronized GPU threads. Buffers in GPU memory are used to store the initial values of equation \eqref{eq-r-matrix}. For the top left chunk of the $R$ matrix (resp. bottom right of $E$), the vertical and horizontal buffers are initialized at $\infty$ (resp. $0$) and the diagonal buffer to $0$ (resp $1$) as detailed in equation \eqref{eq-r-matrix} (resp. \cite{cuturi2017soft}). For the following chunks the horizontal ($B$-wise) buffer stores the upper chunks' last lines, the vertical buffer ($Y$-wise) stores the left chunks' last columns and the diagonal buffer stores the bottom right value of the upper-left chunk. The $E$ matrix is computed similarly, but in the opposite direction.}
    \label{fig:gpu-buffers}
\end{figure}

The $R$ tensor computation takes as input the $B$ and $Y_i's$ time series, and computes each chunk in increasing time direction, as illustrated in Figure~\ref{fig:gpu-chunks}.  The computation of individual values inside each chunk uses thread synchronization schemes similar to those of \cite{zhu2018developing}. Each chunk of the $R$ tensor is memorized into a stack in the CPU memory. The $E$ tensor computation is also performed by chunks in reverse order, and takes as input the $B$ and $Y_i's$ time series and the corresponding $R$ chunk, and produces the $E$ tensor, from which the gradient vector is computed according to equation \eqref{eq:eq-gradient}.

The values of lines, columns and corners between adjacent chunks are transferred between three buffers $hbuf$, $vbuf$ and $dbuf$ of respective shapes $(T, m_b, 1)$, $(T, m_Y, 1)$ and $(T, \lceil m_b/l_b \rceil, \lceil m_Y/l_Y \rceil)$ that are kept in GPU memory, in order to compute the first values of the next chunks, as illustrated in Figure~\ref{fig:gpu-buffers}. The individual values of the $R$ and $E$ matrices are computed in the classical dynamic programming manner described in \cite{cuturi2017soft} as well as \cite{zhu2018developing} and \cite{schmidtCuDTWUltraFastDynamic2020}.

Finally, the division of matrices in chunks helps leverage the stream mechanism of cuda-enabled GPUs. Two streams are used: one stream is in charge of the data transfer between the main memory and the GPU memory, while the other stream is in charge of kernel invocation, i.e. computation and allocation of shared and local memory.

The overall data flow of the proposed computation is as follows:
\begin{enumerate}
    \item The $B$ and $Y_i's$ time series are sent to the GPU memory and the buffers are initialized to hold the initial conditions of equation \eqref{eq-r-matrix} (i.e. $\infty$ in $hbuf$ and $vbuf$, and $0$ in $dbuf$).
    \item Every chunk of the $R$ tensor is computed on GPU, using equation \eqref{eq-r-matrix} adapted to use buffers for initial values. After each chunk is computed, the buffers are then updated as of Figure~\ref{fig:gpu-buffers}, and the chunk is sent back to a LIFO stack in main memory.
    \item A $B$-shaped vector is allocated in GPU memory to store the gradient vector $\nabla_B \sum_i \textit{Soft-DTW}^\gamma(B, Y_i)$ and initialized to $0$. The buffers are filled with the initial conditions for computing $E$ (i.e. $0$ for $hbuf$ and $vbuf$, and $1$ for $dbuf$).
    \item Fourth, the $R$ memory chunks are sent back in reverse order to GPU memory, and their corresponding $E$ matrix chunk is computed. After each $E$ chunk is computed, the gradient vector is incremented according to equation \eqref{eq:eq-gradient}, and both $R$ and $E$ chunks are discarded. 
    \item After all $R$ and $E$ chunks have been processed, the gradient vector is returned in the main memory.
\end{enumerate}

When computing the barycenter of signals with above $10^4$ time steps, our implementation can reduce the computation time by up to $95\%$. An evaluation of the computation time reduction is done in subsection \ref{Step_2_bary}.

The \wetsand prototype computation is obtained by applying the GPU-based Soft-DTW barycenter computation to the subset $C_{train}$ of $C_{traj}$ (cf. Section \ref{modeling}), $C_{traj}$ being a set of cycle trajectories obtained in the segmentation Step 1. This makes it possible to handle signals of varying length and irregularly sampled due to missing data. Assuming that data gaps are randomly scattered throughout the cycles, a sufficiently large $C_{train}$ set should let the Soft-DTW barycenter $B$ of $C_{train}$ be a realistic prototype of normal cycles.  

\subsection{Online outlier detection} \label{Anomaly_detection}
The third and last function of \wetsand consists in using the barycenter $B$ as a reference to detect outlier time series online. $B$ is used to compute a similarity score, inspired by \cite{DsDTW_Jiang}, and defined in Equation \ref{score_dtw}) for each trajectory in the test set $C_{test}$, issued from $C_{traj}$ (cf. Section \ref{modeling}). A threshold is chosen to decide whether a cycle is normal of abnormal.

\begin{definition} [Similarity score]
Given a barycenter $B$, a cycle $C_k$, and their optimal alignment path $\pi$, the \emph{similarity score} $d_{score}$ of $C_k$ is a normalization of the DTW distance that compensates for the fact that trajectories have different lengths: 
    \begin{equation} \label{score_dtw}
    d_{\text {score}}(B, C_{k}) = \frac {\text {DTW}(B, C_k)}{|\pi|}
    \end{equation}
\end{definition}

Here, the DTW score is preferred to the Soft-DTW score to evaluate trajectories because it is deterministic and sometimes more accurate in corner cases. It is also invariant to time shifts while Soft-DTW is not entirely.

There exist many methods to select a threshold, and choosing one heavily depends on the distribution of $d_\text{score}$ in the population. Section \ref{exp_outlier_detection} evaluates different methods applied to this particular case. Ultimately, choosing an appropriate threshold is out of the scope of \wetsand and left to the user.

\section{\wetsand applied to industrial robots} \label{Evalu_approach}

In this section, \wetsand is evaluated on the case study described in Section~\ref{Motivating_usecase}. 
For all our experiments, the Python programming language on a cloud instance with a NVIDIA A10G GPU, an 8-Core AMD EPYC 7R32 CPU and a limit of 31GB of memory has been used.
When relevant, the proposed GPU-based Soft-DTW barycenter implementation is compared to that of the TS Learn library in terms of speed and results. It is a cross-platform software package for Python 3.5+ \cite{TSLearn_Tavenard} that provides efficient implementations for Soft-DTW and Dynamic Time Warping Barycenter Averaging (DBA) (see \cite{petitjean2011global}) algorithms. This library is itself based on scikit-learn \cite{scikit-learn}. 

The datasets are described in Section \ref{dataset}, and the assessment of the results, based on both quantitative and qualitative evaluations, is described in Section \ref{Eval_metrics}.

\subsection{Datasets production} \label{dataset}

The task of a robotic arm is defined by a script, that specifies a number of way points in the robot's task space. The robot's controller computes the corresponding target joint trajectory in order to reach these way points. When the robotic arm moves, its internal encoders read the actual joint trajectory. These values - way points and joint positions - are collected from the controller into a data storage and form $Y \in \mathbb{R}^{d \times N}$, where $d=9$.

\begin{figure}
    \centering
    \includegraphics[width=\linewidth]{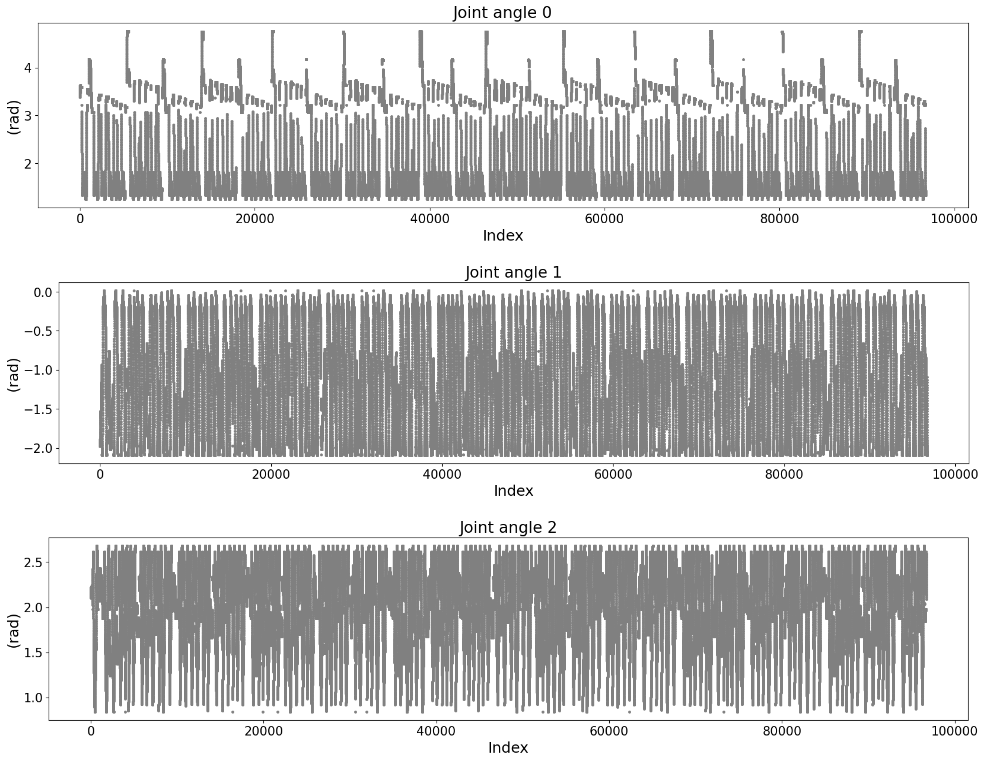}
    \caption{Robot angular position monitored during several task cycles, by internal sensors in joint 0 (base), joint 1 (elbow) and joint 2 (shoulder).}
    \label{fig: raw_data}
\end{figure}

For experimentation purposes, a robot, named ''robot A'', whose task is to fill containers has been selected. This robot fulfills the working hypotheses, due to the cyclic aspect of its movements on each of its monitored joints (cf. Figure~\ref{fig: raw_data}). The monitoring data also features numerous and large gaps.

The dataset for the segmentation step defined in subsection~\ref{Sub_DTW} is composed of the robot movement captured from its first three joints (base is joint 0, elbow is joint 1 and shoulder is joint 2) during 9 hours with no particular incident documented.

The segmentation results in a pool of 217 cycles \(C_{train_{1}},...,C_{train_{217}} \) represented by three matrices, one per joint, of 1 row and respectively \(m_1,...,m_{217}\) columns.
Several cycles - the number varies across experiments - are randomly selected to create a training set $C_{train}$ for the prototype computation step.
Then, the evaluation of the results and the outlier detection step is done on a test set \(C_{test}\) of 190 randomly selected cycles among the rest.
As explained in subsection~\ref{modeling}, each cycle has different length; to respect the requirements of Soft-DTW's implementation, all the cycle are padded with their respective last values to the length of the longest.

Another monitored robot, named ``robot B'', is used to provide the dataset for the experiments of~\ref{Segmentation}. The task of robot B is to depalletize devices and place them into a machine. Its movement is much more complex and less smooth that that of robot A; due to the industrial setup of the data collection, this behavior increases the amount of missing data. The dataset acquired on robot B represents the worst case scenario to test the limitations of our approach.

\subsection{Evaluation means and metrics} \label{Eval_metrics} 

\begin{figure}
    \centering
    \includegraphics[width=\linewidth]{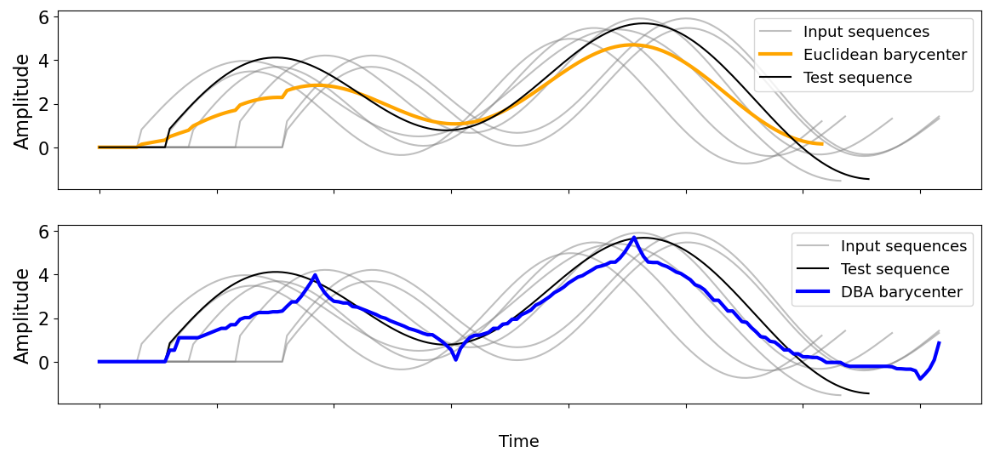}
    \caption{Illustration of the computation of Euclidean mean (above) and DBA barycenter (below) from an input set in grey. DBA tends to optimise better the sum of DTW distances, but is less interpretable.}
    \label{fig: example_barycenters}
\end{figure}

Abnormal cycles detected by \wetsand are meant to be analysed by a maintenance operator. This is why strong emphasis is placed on \emph{visual} inspection of the results for the segmentation and the prototype computation steps. For instance, Figure \ref{fig: example_barycenters} illustrates that a barycenter can easily be assessed visually. It is clear that the DBA barycenter, despite being excellent at minimizing elastic distances, looses the shape of the input signals, thus making its inner working confusing to a non-expert user. 

The intraclass inertia with respect to the DTW distance is also used (the lowest value, the best). In Figure~\ref{fig: example_barycenters} the intraclass inertia for the Euclidean mean is $354.5$ and the one for the DBA barycenter is $97.2$, which shows that DBA is better suited to compare warped trajectories than Euclidean distance.

For computation times, comparison is done with the equivalent algorithms in the TS Learn library, as they provides high quality, efficient implementations.

For evaluating the overall anomaly detection performance, the whole $C_{test}$ set has been manually annotated into normal and abnormal cycles. While the whole \wetsand approach is completely unsupervised, this allows us to compare the results of \wetsand and the other approaches with the annotated ground truth. Confusion matrices and $F_1$ score are used to evaluate various means to set the detection threshold, as well as the global \wetsand approach.

\section{Experiments and results} \label{Experiments}

This section provides a detailed analysis of the three steps of  \wetsand and evaluates their performances via the metrics described in subsection \ref{Eval_metrics}.

\subsection{Evaluation of Step 1 -- Segmentation into cycles} \label{Segmentation}

The aim of the first experiment is to partition a long dataset of monitored robot movement into cycle trajectories by applying Sub-DTW through a sliding-window. While \cite{lacoquelle2022deviation} demonstrates a similar function, their reference cycle $X$ is extracted from a simulator, which is better to avoid, given that a simulator is difficult to build or obtain from the robot manufacturer. Instead, an operator is asked to limit one cycle even if it is distorted and has data gaps, and use this cycle  as the reference.

\begin{figure} 
\centering
\small 
\includegraphics[width=\linewidth]{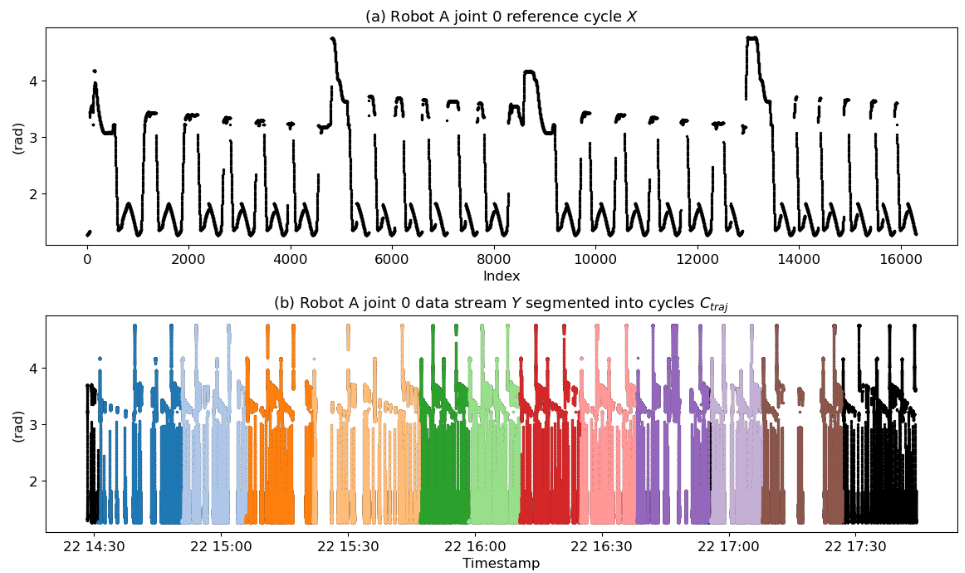}

\includegraphics[width=\linewidth]{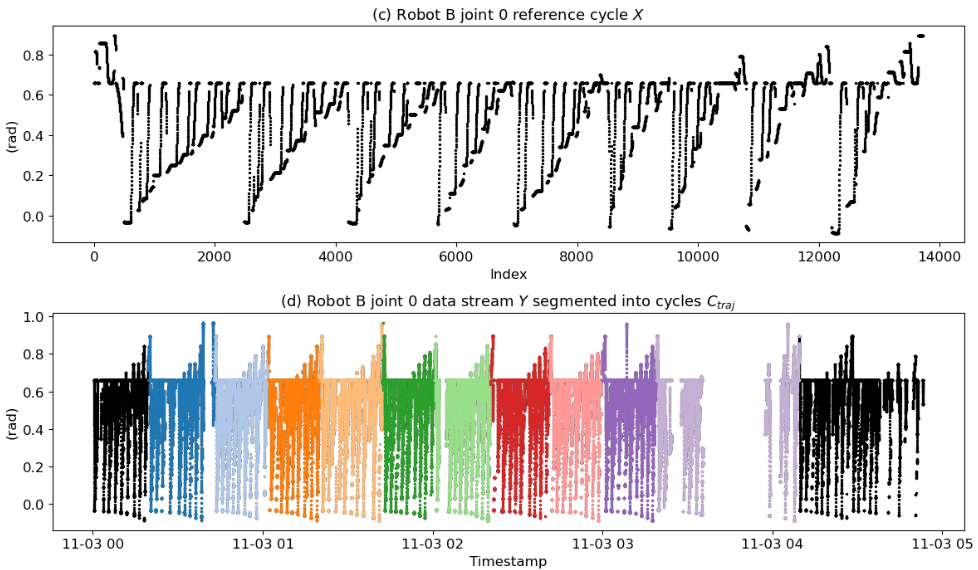}

\caption{Segmentation of the movement data of two robot datasets into cycles. Above, the reference cycle (a) and segmented data stream (b) for robot A. Below, the reference cycle (c) and the segmented data stream (d) for robot B. In (b) and (d) black represents the samples that have not been associated with a complete cycle.}
\label{fig: segmentation_j0}
\end{figure}

Figure~\ref{fig: segmentation_j0} reports the result of such segmentation in which each limited cycle has different color. Qualitative visual evaluation was conducted on each signal and tends to shows that the segmentation algorithm works well and identifies the real bounds for the task cycles in the data stream.

The computation time is also satisfactory. The algorithm scales up to $Y$ streams of around $6.10^5$ time steps. A single segmentation step takes around 3 seconds in the case of robot A that has a reference cycle of $16\,300$ time steps.

As described in section~\ref{dataset} the dataset consists of 9 synchronous time series for each robot joint and 3 Cartesian coordinates, with synchronous data gaps. 
We found that using all of the robot's joint and task space variables all together as input to Sub-DTW enhances the accuracy of the segmentation. Thus, cycles are successfully characterized despite the poor quality of both the query and the search sequence.

\subsection{Evaluation of Step 2 -- Prototype computation} \label{Step_2_bary}

The second experiment aims at assessing the quality of the GPU-based Soft-DTW barycenter implementation presented in Section \ref{Soft_DTW}. In this experiment, each joint is analyzed separately. The reason for this is that an anomaly due to an internal failure may only be visible on one joint. Hence, there is a potential for the anomaly to become less noticeable when blended with the surrounding normal joints.

The desired qualitative properties of the barycenter are:
\begin{itemize}
    \item to accurately represent the true movements of the robot during one cycle;
    \item to ensure the absence of any data gaps, despite the presence of numerous and sizable gaps in the original cycles of the training set $C_{train}$.
\end{itemize}
The barycenter computed by the GPU-based Soft-DTW implementation is also visually compared with the Euclidean mean, and the DBA barycenter provided by TS Learn.  This allows one to assess the accuracy with which each barycenter faithfully represents a genuine cycle.

\begin{table*}
\centering
\caption{Hyper-parameters used in \wetsand.}
\label{tab:params_soft}
\begin{tabular}{|p{0.15\linewidth}|p{0.5\linewidth}|p{0.25\linewidth}|}
\hline
\textbf{Parameter} & \textbf{Description} & \textbf{Value} \\ \hline
Gamma & The DTW softness parameter (used in softmin) & 1 (default) \\ \hline
Bandwidth & The Sakoe-Chiba bandwidth, as a fraction of the longest signal length & 0.6 \\ \hline
Init  & Initial barycenter to start from for the optimization process & Euclidean barycenter (default) \\ \hline
Method  & Optimization method & L-BFGS (default) \\ \hline
Max iter  & (scikit-learn) maximum number of gradient descent updates & 40 \\ \hline
Max fun  & (scikit-learn) maximum number of times soft-dtw barycenter is computed & 200 \\ \hline
G tol & (scikit-learn) gradient-based termination parameter for the L-BFGS algorithm & $1.e^{-8}$  \\ \hline
Tolerance & (scikit-learn) distance-based termination parameter for the L-BFGS algorithm & $1.e^{-5}$ \\ \hline
Chunk shape &
  Tuple $(nx, ny, tx, ty)$ where $nx$ (resp $ny$) is the number of signals from $X$ (resp $Y$) and tx (resp ty) is the number of timesteps from x (resp y) &
  $(1, |C_{traj}|, 2^9, 2^{10})$ \hbox{(default)} \\ \hline
\end{tabular}
\end{table*}

\begin{table*}
\centering
\caption{Comparison of Euclidean mean, DBA and the Soft-DTW barycenters \\
w.r.t their ability to synthesize a proper robot trajectory \label{tab:results_comp_bary}}
\begin{tabular}{|p{19mm}|c|c|c|c|c|c|}
\hline
\multirow{3}{*}{\textbf{Algorithm}} &
  \multicolumn{2}{c|}{\textbf{Joint 0}} &
  \multicolumn{2}{c|}{\textbf{Joint 1}} &
  \multicolumn{2}{c|}{\textbf{Joint 2}} \\ \cline{2-7} 
 &
  Intraclass & Mean &
  Intraclass & Mean &
  Intraclass & Mean \\
 &
  inertia & $d_{score}$ &
  inertia & $d_{score}$ &
  inertia & $d_{score}$ \\
  \hline
  Euclidean mean &
  \multicolumn{1}{l|}{30440} &
  0.197 &
  \multicolumn{1}{l|}{25725} &
  0.159 &
  \multicolumn{1}{l|}{15979} &
  0.103 \\ \hline DBA &
  \multicolumn{1}{l|}{3808} &
  0.028 &
  \multicolumn{1}{l|}{5411} &
  0.032 &
  \multicolumn{1}{l|}{3413} &
  \textbf{0.023} \\ \hline GPU-based Soft-DTW &
  \multicolumn{1}{l|}{\textbf{1516}} &
  \textbf{0.017} &
  \multicolumn{1}{l|}{\textbf{2653}} &
  \textbf{0.025} &
  \multicolumn{1}{l|}{\textbf{2484}} &
  \textbf{0.023} \\ \hline
\end{tabular}
\end{table*}

\begin{figure} 
\centering
\includegraphics[width=\linewidth]{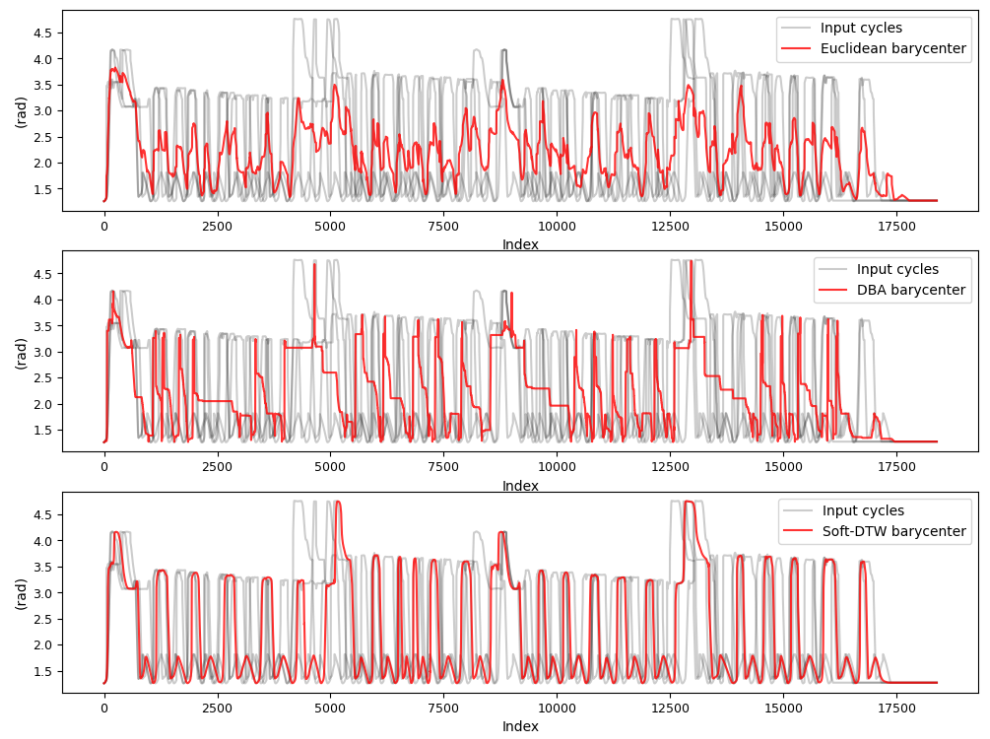}
\caption{Comparison of barycenters obtained by Euclidean mean (top), by DBA (middle) and by Soft-DTW 
(bottom) on joint 0's angular position. In grey, the 5 input time series and in red the barycenter.}
\label{fig: example_3barycenters}
\end{figure}

As illustrated in Figure \ref{fig: example_3barycenters}, the Soft-DTW barycenter has the qualitative properties, and it obviously outperforms the Euclidean mean and the DBA barycenter in terms of representation faithfulness. Interestingly, this result has been obtained with only 5 trajectories in the training set. The hyper parameters used for the Soft-DTW computation are listed in Table \ref{tab:params_soft}.

The same conclusion is obtained with quantitative metrics, as provided in Table~\ref{tab:results_comp_bary}. On each of the three joints, Soft-DTW has the lowest $intraclass$ $inertia$ and mean $d_{score}$ compared to the Euclidean mean and the DBA barycenter.

\begin{table}
\caption{{Soft-DTW barycenter computation time with TS Learn and \wetsand implementations on signals with 18385 time steps}.\label{tab: Computation_time}}
\centering
\begin{tabular}{|p{0.4\linewidth}|c|c|c|}
    \hline
    Number of cycles in the training set $C_{train}$ & 5 & 10 & 25 \\ \hline
    \mbox{TSLearn Soft-DTW}\par \mbox{computation time (minutes)} & 777 & 1217 & 3933 \\ \hline
    \mbox{GPU Soft-DTW}\par \mbox{computation time (minutes)} & \textbf{40.3} & \textbf{64} & \textbf{128}  \\ \hline
\end{tabular}
\end{table}

A final comparison to the TS Learn package is done regarding the computation time. The barycenter computation is achieved on the three joints of robot A with different input training set sizes via both implementations. The results are displayed in Table~\ref{tab: Computation_time}. 
The relevance of this comparison is guaranteed by the fact that the barycenters computed by each implementation are alike - indeed the $d_\text{score}$'s order of magnitude between both of them is $10^{-3}$ which is extremely low for normalized signals of length 18385.

\subsection{Evaluation of Step 3 -- Online outlier detection} \label{exp_outlier_detection}

The ultimate target of \wetsand is the anomaly detection step that aims at labeling cycles as "normal" and "abnormal" according to their similarity with their respective joint's barycenter. Step 3 takes as input the barycenter computed in the previous experiment (cf. Section \ref{Step_2_bary}), and the test set $C_{test}$. It computes the $d_\text{score}$ as of equation \eqref{score_dtw} for each sample of $C_{test}$. For the same reasons as for Step 2, the $d_\text{score}$ is computed independently for each joint.

The evaluation of Step 3 is divided into three parts. First, the distribution of the $d_\text{score}$ values is inspected with box plots to assess the feasibility to set a detection threshold. Second, the expert annotations on the test dataset (described earlier in Section~\ref{Eval_metrics}) are used to establish the confusion matrix for different threshold values. Third, the results are compared with deep-learning approaches using several variants of autoencoders.

\subsubsection{Setting the detection threshold}

\begin{figure*} 
\centering
\small 
\begin{tabular}{ccc}
Robot A, joint 0 & Robot A joint 1 & Robot A joint 2 \\
\includegraphics[width=0.3\linewidth]{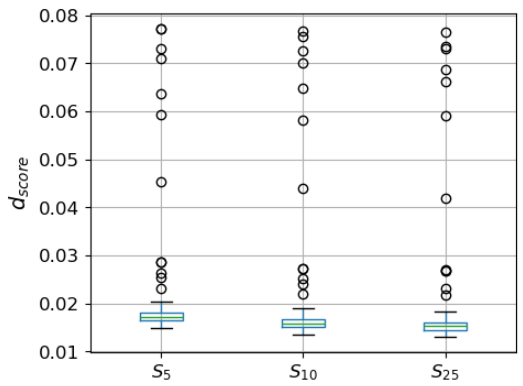} &
\includegraphics[width=0.3\linewidth]{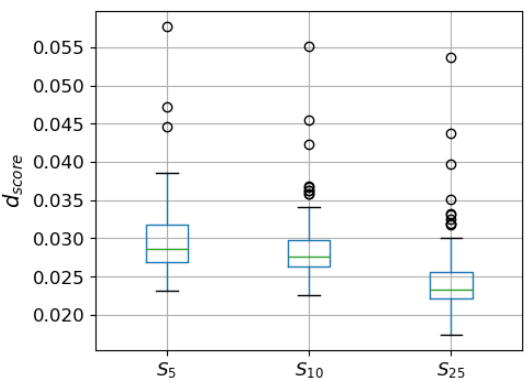} &
\includegraphics[width=0.3\linewidth]{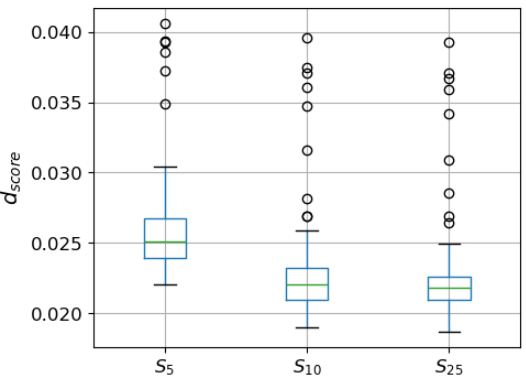}
\end{tabular}

\caption{$d_{score}$ distribution boxplots for 3 joints of robot A with 5, 10 and 25 cycles in the training set $C_{train}$, i.e. $S_5$, $S_{10}$, and $S_{25}$ sizes for $C_{train}$ respectively.}
\label{fig: dscore_boxes}
\end{figure*}

The boxplots of the distributions of $d_\text{score}$ are depicted in Figure~\ref{fig: dscore_boxes} for different sizes of the training set $C_{train}$: $S_{5}$ for 5 cycles, $S_{10}$ for 10 cycles, and $S_{25}$ for 25 cycles. The illustrated result is the mean of 10 randomly sampled training sets. The $d_\text{score}$ clearly shows a sparse tail distribution for all joints and configurations, which makes a threshold based approach suitable. 

Two possible ways to set the detection threshold are considered. First, the threshold is set at two standard deviations of the mean - but one could also set the threshold at three standard deviations. It is then called the \textit{$2\sigma$ threshold}. Second, the threshold is based on the boxplot. It is then called the \textit{boxplot threshold} and set to \(Q_{max} = q_3 + 1.5*|q_3 - q_1| \), where $q_1$ and $q_3$ are the first and the third quartiles of the distribution respectively (cf. Figure \ref{fig: dscore_boxes}).

\subsubsection{Evaluation against the expert annotations}

\begin{figure} 
\centering
\includegraphics[width=\linewidth]{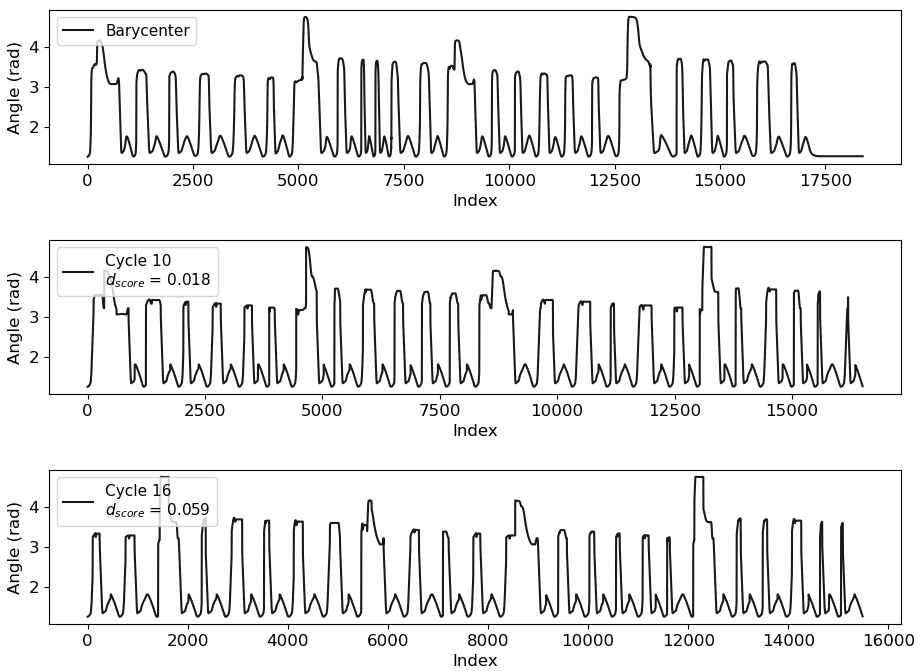}
\caption{The Soft-DTW barycenter for joint 0 of robot A (top), a normal (middle), and an abnormal cycle (bottom).}
\label{fig: bary_test_traj}
\end{figure}

The remainder of this section only considers the scores computed with a barycenter obtained with a training test of $5$ cycles for conciseness. The relationship between training set size and $d_\text{score}$ distribution is further discussed in Section~\ref{discussion}. The following evaluations of the anomaly detection step rely on expert annotations of the test dataset, as illustrated in Figure~\ref{fig: bary_test_traj}, which depicts the Soft-DTW barycenter for joint 0 of robot A, a cycle labeled ``normal'', and a cycle correctly labeled ``abnormal''. Similarity can be assessed by counting the number of peaks - which illustrate the joint going back and forth - and judging their amplitude.

 The resulting confusion matrices are shown in Table \ref{tab:WETSAND_confusion}. These are further analyzed through the $F_2$ score, a popular metric in Machine Learning. In the considered industrial context, the drawbacks of having false negatives (undetected anomalies) outweighs those of having false positives (spurious anomalies). In other words, \textit{recall} is favored against \textit{precision}, which means obtaining the highest $F_2$ score. 
Given that the \(2\sigma\) threshold results in \(F_2 = 0.69\) and the boxplot threshold results in \(F_2 = 0.92\), the threshold is set at the boxplot threshold for the experiments that follow.

\begin{table}
\centering
\caption{Confusion matrix between ground truth and \wetsand\\ prediction on joint 0 of robot A using a $2\sigma$ and a boxplot threshold.}
\label{tab:WETSAND_confusion}
\begin{tabular}{@{}c|c|c|c|c|c|c|}
\multicolumn{2}{l}{} & \multicolumn{2}{c}{\wetsand} & \multicolumn{1}{l}{} & \multicolumn{2}{c}{\wetsand} \\
\multicolumn{2}{l}{} & \multicolumn{2}{c}{$2\sigma$ threshold} & \multicolumn{1}{l}{} & \multicolumn{2}{c}{boxplot threshold} \\
\cline{3-4}\cline{6-7}
\multicolumn{2}{l|}{} & Anomaly & Normal & & Anomaly & Normal \\
\cline{2-4}\cline{6-7}
Ground & Anomaly & 7 & 4 & & 11 & 1 \\
\cline{2-4}\cline{6-7}
truth & Normal & 0 & 179 & & 1 & 177 \\
\cline{2-4}\cline{6-7}
\end{tabular}
\end{table}

\subsubsection{Evaluation against deep autoencoders}
The performance of \wetsand is compared against that of a Convolutional AutoEncoder (CAE).
CAE are often used for anomaly detection in time series through the use of 1D convolutional layers. Autoencoder-based anomaly detection consists in training the autoencoder on a set that contains only (or mostly) normal individuals. Under this training protocol, it is assumed that the CAE  reconstructs new normal individuals - which are similar to those used for training - better than abnormal ones. Then, the reconstruction error that is the loss function used to train the autoencoder naturally becomes a $d_\text{score}$ for anomaly detection.

While autoencoders can learn extremely elaborate datasets, they usually require massive amounts of training data. This is incompatible with the industrial setting at hand, where robots are daily recalibrated or reassigned to a different task. For a fair comparison, autoencoders are trained with the same number of cycles as the Soft-DTW barycenter, i.e. on a training set $C_{train}$ of 5 trajectories. As a consequence, architectures that can compact signals of $18\,000$ time steps into a small latent space with few parameters are aimed for.

\begin{figure} 
\begin{center}
\includegraphics[width=0.8\linewidth]{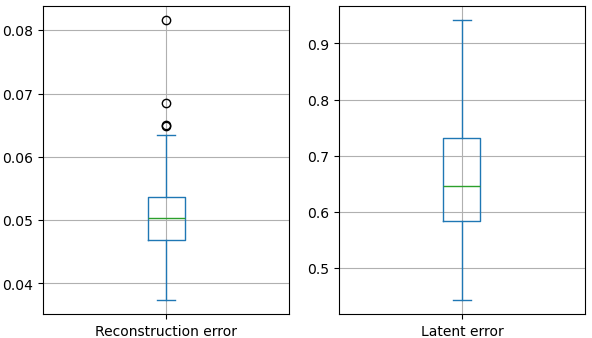}
\end{center}
\caption{Reconstruction error (left) and error computed on the latent space (right) of the baseline CAE on a test set of 190 trajectories.}
\label{fig: loss_scores_AE}
\end{figure}

To that end, investigations were conducted to find the best CAE architecture. The time series are first last-value-padded to 24576 time steps, which is the maximal number of time steps of the cycles in the training set. A first convolutional layer with layer kernel size 32, stride 32, and 16 output channels is followed by a rectified linear unit (ReLU) layer used for non linearity. A second convolutional layer uses kernel size 32, stride 32, and 16 output channels, thus yielding a latent space of dimension $5*16*24 = 1\,920$, and a total of $17\,457$ trainable parameters. Decoding is performed with two 1D transpose convolutional layers with the same parameters, separated by a ReLU layer. Fine tuning is computed with the adaptative moment estimation (ADAM) algorithm with a learning rate of $10^{-3}$. The parameters were chosen to minimize the loss score during the training stage, while minimizing the number of parameters and therefore, the training duration. 
The reconstruction error distribution is depicted on the left of Figure~\ref{fig: loss_scores_AE}. Setting the boxplot anomaly threshold yields the confusion matrix depicted in Table~\ref{tab:CAE_confusion}, and an $F_2$ score of $0.10$, which is a quite poor result.

\begin{table}
\centering
\caption{Confusion matrix for anomaly detection using a Convolutional Autoencoder \label{tab:CAE_confusion}}
\begin{tabular}{c|c|c|c|}
\multicolumn{2}{l}{}  & \multicolumn{2}{c}{Convolutional} \\
\multicolumn{2}{l}{}  & \multicolumn{2}{c}{autoencoder} \\
\cline{3-4} 
\multicolumn{2}{l|}{} &  Anomaly & Normal \\
\cline{2-4}
Ground & Anomaly & 1 & 11 \\
\cline{2-4} 
truth & Normal & 3 & 175 \\
\cline{2-4} 
\end{tabular}
\end{table}

Variants of the CAE model have been tested to improve the anomaly detection performance, such as adding Attention layers, which reduces the total number of parameters of the model.
In particular, a test was done with a CAE whose encoder consists of a first layer of kernel size 32, stride 32, 8 output channels, feeding second layer of 8 channel multihead attention (in all query, key and value inputs) followed by a third convolutional layer of kernel size 16, stride 16 and 8 output channels, with two ReLU layers in between. The decoder is symmetric, with the same attention layer and convolutions being replaced with transpose convolutions.
This CAE has a latent space dimension of $5*8*48 = 1\,920$, but only $3\,161$ trainable parameters.
It yields an $F_2$ score of 0.23, which is better that the previous CAE but is still much lower than the anomaly detection performance with \wetsand. Other autoencoder variants, such as variational autoencoders, recurrent layers, or using the latent space distance as a detection threshold (an idea presented in \cite{akcay2019ganomaly}), did not yield better results.

\begin{figure} 
\includegraphics[width=\linewidth]{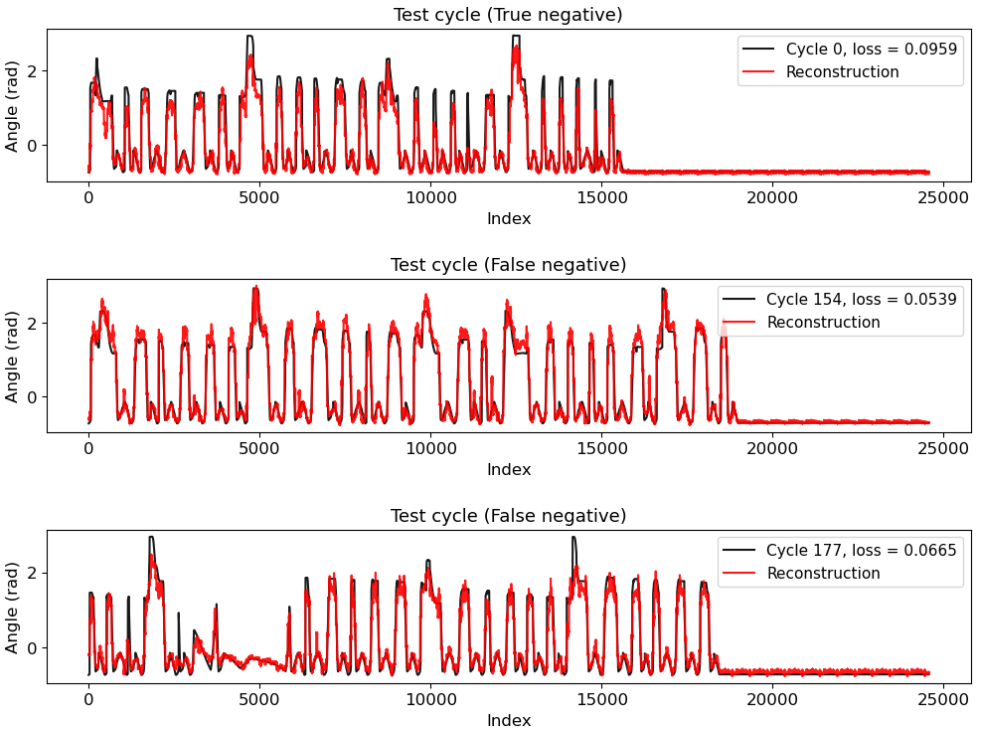}
\caption{Examples of reconstructed cycles (in red) with the Attention-CAE compared to the input cycle (in black). Every cycle has a low $d_\text{score}$, even though the middle and lower ones bear anomalies.}
\label{fig: result_attention_cae}
\end{figure}

An explanation for the poor performance of autoencoders is illustrated in Figure~\ref{fig: result_attention_cae}. Even with a low number of parameters, the autoencoders reconstruct normal and abnormal cycles equally well. Increasing the number of network trainable parameters only worsens the problem. Conversely, decreasing the number of trainable parameters to the limit of underfitting only makes the autoencoder equally bad at reconstructing normal and abnormal cycles. Using recurrent layers or larger training sets also yields similar results. 

An interpretation is that the nature of targeted anomalies does not bode well with autoencoders. Targeted anomalies rarely cause out of bounds signal points, but are rather characterised by a missing or an undesired pattern, e.g. longer or shorter than usual. An autoencoder that has learned the nominal robot pattern can easily reconstruct a cycle with an additional (or missing) pattern with the same shape. This explains why autoencoders generalize too well to abnormal cycles in the presented use case, and fail at detecting them. In contrast, elastic distances such as DTW used in \wetsand must match each pattern to a reference,  exhibiting high sensitivity to the absence or inclusion of undesired patterns, which perfectly fits the need of targeted anomalies. Another argument against CAE in this application is that the tuning of depth and hyperparameters is long, and it is specific to each type of time series. It is unlikely that in the industrial context, one would fine tune a model for each robot and each robots' joints. 

\subsection{Discussion on hyperparameter tuning} \label{discussion}
In this subsection, the impact of several hyperparameters of the Soft-DTW algorithm on the barycenter computation are evaluated. We report on the tuning of 2 hyperparameters:
\begin{itemize}
    \item barycenter initial value;
    \item training set size;
\end{itemize}

\subsubsection{Barycenter initial value tuning experiment}
 3 initialization methods for the gradient descent phase of the Soft-DTW barycenter are compared: zero, random, and euclidean mean. It has been observed that initializing the algorithm with an array of zeros makes gradient descent $10\%$ slower than when randomly initialized. However, the produced barycenter is more similar to the normal trajectory. Initializing with the euclidean mean of the training set makes gradient descent $8\%$ faster than zero-initialization, and yields an even better barycenter.

\subsubsection{Training set size tuning experiment}
This experiment reports on the influence of the size of the training set $C_{train}$ on the barycenter computation. The intuitive idea driving this experiment is that a higher number of cycles in the training set will lead to a more representative barycenter. Consequently, this is expected to result in a higher similarity score between the barycenter and the normal test cycles. Here, the GPU-based implementation is run with the parameters described in Table \ref{tab:params_soft}. Three training sets are tested: the subsets $S_5$, $S_{10}$, $S_{25}$, which are randomly selected from the cycle pool, such that \(|S_5|=5, |S_{10}|=10, |S_{25}|=25\), and \(S_5 \subset S_{10} \subset S_{25}\). Besides, the experiment is done independently on the three joints of robot A and each of them is repeated ten times for statistical accuracy; then, the mean result for each sample is recorded. The $d_{score}$ between the computed barycenters and the test set according to the training set size on each joint are presented on Figures \ref{fig: dscore_boxes}. A lower median value for $S_{25}$ on each experiment demonstrates the validity of the intuition. It is also noticeable in Table~\ref{tab: F2_scores_exp3} that, depending on the signal's shape, the training set size has an impact on the sensitivity of the $d_{score}$ to outliers. While joint 0's barycenter is satisfying with $S_5$ as a training set, the anomaly detection is improved with a barycenter computed with a larger training set for joints 1 and 2.

\begin{table}
\caption{{$F_2$ scores on the test set of 190 cycles, with \wetsand implementation on Robot A's three joints}.\label{tab: F2_scores_exp3}}
\centering
\begin{tabular}{|p{0.4\linewidth}|c|c|c|}
    \hline
    Number of cycles in the training set $C_{train}$ & 5 & 10 & 25 \\ \hline
    \mbox{$F_2$ score} \mbox{Robot A Joint 0} & \textbf{0.92} & \textbf{0.92} & \textbf{0.92} \\ \hline
    \mbox{$F_2$ score} \mbox{Robot A Joint 1} & 0.42 & 0.80 & \textbf{0.95}  \\ \hline
    \mbox{$F_2$ score} \mbox{Robot A Joint 2} & 0.79 & \textbf{0.97} & \textbf{0.97}  \\ \hline
\end{tabular}
\end{table}





\section{Conclusion} \label{Conclu}
In this paper, a frugal and efficient anomaly detection approach named  \wetsand is introduced.  It is designed to handle cyclostationary time series with irregular sampling and distortion in the time dimension. \wetsand segments a large quasiperiodic time series into quasiperiodic cycles, using as input a (possibly corrupted) cycle. It then automatically computes a representative prototype of the normal cycles using an efficient and new GPU-based implementation of the Soft-DTW barycenter, specifically tailored for long time series. Outlier time series detection is then performed using the DTW distance between the computed barycenter and upcoming data. 

\wetsand's efficiency is evaluated with several experiments performed on real data from industrial robots. It is demonstrated that \wetsand scales up to times series in the tens of thousands of time steps, and outperforms autoencoder based approaches on the data set of repetitive robotic arm motions. This is another argument in favor of the assertion developed in~\cite{audibert2022deep} that says that deep neural network methods do not necessarily outperform conventional approaches and should not be used systematically.

\section*{Acknowledgments}
The authors would like to thank Christophe Merle, Head of the AI Group at Vitesco Technologies and ANITI Industrial Coordinator, for sharing his expertise in robotized production lines and for his sound advice.
This work was supported by the AI Interdisciplinary Institute ANITI, funded by
the French program ``Investing for the Future -- PIA3'' under Grant agreement no.
ANR-19-PI3A-0004.

\bibliographystyle{IEEEtran}
\bibliography{ieee}


\section{Biography Section}
 



To be completed after anonymous review

\vfill

\end{document}